%% file: main.tex
\documentclass[10pt,letterpaper]{article}
\usepackage[top=0.85in,left=2.5in,footskip=0.75in]{geometry}

\usepackage{amsmath,amssymb}
\usepackage[square,sort,comma,numbers]{natbib}
\usepackage{changepage}

\usepackage[utf8x]{inputenc}

\usepackage{textcomp,marvosym}
\usepackage{cite}

\usepackage{nameref,hyperref}
\usepackage{times}
\usepackage{float}
\usepackage{mathrsfs}
\usepackage{amsthm}
\usepackage{comment}
\usepackage[center]{caption}
\DeclareMathOperator*{\argmin}{argmin}
\usepackage{booktabs}
\usepackage{graphicx}
\usepackage{capt-of}
\usepackage{booktabs}
\usepackage{varwidth}
\usepackage[T1]{fontenc}

\usepackage{pgf,tikz}
\usetikzlibrary{topaths,calc}
\usetikzlibrary{arrows,automata}
\usetikzlibrary{arrows.meta}

\usepackage{microtype}
\DisableLigatures[f]{encoding = *, family = * }

\usepackage{array}

\newcolumntype{+}{!{\vrule width 2pt}}

\newlength\savedwidth

\raggedright
\newtheorem{example}{Example}
\newtheorem{theorem}{Theorem}
  \newtheorem{lemma}[theorem]{Lemma}
  \newtheorem{proposition}[theorem]{Proposition}
  
  \newtheorem{corollary}[theorem]{Corollary}
  \newtheorem{definition}[theorem]{Definition}

\usepackage[aboveskip=1pt,labelfont=bf,labelsep=period,justification=raggedright,singlelinecheck=off]{caption}

\makeatletter
\renewcommand{\@biblabel}[1]{\quad#1.}
\makeatother

\usepackage{lastpage,fancyhdr,graphicx}
\usepackage{epstopdf}
\fancyheadoffset[L]{2.25in}
\fancyfootoffset[L]{2.25in}
\begin{document}
\vspace*{0.2in}

\begin{flushleft}
{\Large
\textbf\newline{Hypergraph Partitioning using Tensor Eigenvalue Decomposition} 
}
\newline
\\
Deepak Maurya,
Balaraman Ravindran
\\
\bigskip
Computer Science and Engineering, \\
Robert Bosch Centre for Data Science and AI, \\
Indian Institute of Technology Madras, Chennai, India 
\bigskip

*corresponding author: ravi@cse.iitm.ac.in

\end{flushleft}
\section*{Abstract}
Hypergraphs have gained increasing attention in the machine learning community lately due to their superiority over graphs in capturing \textit{super-dyadic} interactions among entities. In this work, we propose a novel approach for the partitioning of $k$-uniform hypergraphs. Most of the existing methods work by reducing the hypergraph to a graph followed by applying standard graph partitioning algorithms. The reduction step restricts the algorithms to capturing only some weighted pairwise interactions and hence loses essential information about the original hypergraph. 

We overcome this issue by utilizing the \textit{tensor}-based representation of hypergraphs, which enables us to capture actual super-dyadic interactions. We prove that the hypergraph to graph reduction is a special case of tensor contraction. We extend the notion of minimum ratio-cut and normalized-cut from graphs to hypergraphs and show the relaxed optimization problem is equivalent to tensor eigenvalue decomposition. This novel formulation also enables us to capture different ways of cutting a hyperedge, unlike the existing reduction approaches. We propose a hypergraph partitioning algorithm inspired from spectral graph theory that can accommodate this notion of hyperedge cuts. We also derive a tighter upper bound on the minimum positive eigenvalue of even-order hypergraph Laplacian tensor in terms of its conductance, which is utilized in the partitioning algorithm to approximate the normalized cut. The efficacy of the proposed method is demonstrated numerically on simple hypergraphs. We also show improvement for the min-cut solution on 2-uniform hypergraphs (graphs) over the standard spectral partitioning algorithm.


\input{sections/intro.tex}
\input{sections/prelims.tex}
\input{sections/hy_contr.tex}
\input{sections/prop_algo.tex}

\input{sections/exp.tex}

\section{Conclusions \& Future Work}
\label{sec:conc}
In this work, we propose a hypergraph partitioning algorithm using tensor eigenvalue framework and establish its superiority over existing hypergraph reduction methods. We extend the notion of ratio-cut and normalized cut from graphs to hypergraph and show the equivalence of relaxed optimization problem to a tensor eigenvalue problem. Further, we derive a tighter upper bound for the approximation of Normalized-cut problem. We also reduce the computation time drastically for Fiedler vector of Laplacian tensor of hypergraph. The future directions of this work is along the lines of similar analysis for non-uniform and directed hypergraphs. 

\section*{Supporting information}
\appendix

\input{sections/appen_proof.tex}

\input{sections/appen_exmpl.tex}

\bibliography{main}            
\end{document}

%% file: sections/intro.tex
\section{Introduction}
In machine learning, interacting systems are often modeled as graphs. In graph modeling, an interacting object is represented as a node, and an edge captures the interaction between a pair of objects. A conventional approach is to quantify the extent of interaction by associating a positive \emph{weight} to the corresponding edge.
This graph formulation is further utilized for various standard machine learning applications like clustering \citep{dhillon2004kernel} and semi-supervised learning \citep{wang2016scalable}. While a graph representation is limited to capturing only \emph{pairwise} interaction, many real-world systems may involve interactions that may be more complex than the simple pairwise formulation \citep{zhou2005beyond}. For instance, a collaboration network may involve agents interacting at a group level (also called super-dyadic interactions), which can not be captured by modeling the system as a graph.

Recently, \textit{hypergraphs} have been used to represent and analyze such complex {\it super-dyadic} relationships. Hypergraphs are generalizations of graphs where an edge could potentially connect multiple nodes. These edges are commonly referred to as \textit{hyperedges}. A $k$-uniform hypergraph refers to the case when all hyperedges are constrained to contain exactly $k$ nodes. Hypergraph partitioning has been used in a variety of applications in several domains, such as VLSI placement \citep{karypis1999multilevel}, object segmentation in videos \citep{huang2009video}, and citation networks \citep{zhou2007learning}.

Existing hypergraph modeling frameworks can be classified into two paradigms, based on whether they reduce the hypergraph to a graph explicitly \citep{paper:agarwal_2006}  or implicitly. These reduction based approaches are quite popular in the machine learning community due to the scalability to large datasets \citep{li2013news, gao20123}, and provable performance guarantees of graph-based algorithms \citep{ghoshdastidar2015provable}. Thus most of the existing approaches make use of hypergraph reduction to utilize standard graph-based algorithms, which defeats the motivation behind using hypergraphs. As graphs are limited to capture only dyadic interactions, the reduction-based approaches fail to model the desired super-dyadic relationships. 

An exciting aspect of hypergraph partitioning is that a hyperedge can be cut in multiple ways, unlike the case of an edge in graphs. The nodes in a hyperedge can be split in different ways. Most of the existing reduction based partitioning methods do not differentiate these multiple configurations and penalize them equally, which, ideally, should not be the case. In fact, Ihler et al. \citep{ihler1993modeling} show that the reduction-based approaches can not model a hypergraph cut, i.e., the complete removal of a hyperedge from a given hypergraph. 

On the other hand, tensors have gained increasing attention for modeling hypergraphs, primarily in the mathematics community. For instance, Hu et al. \citep{hu2014eigenvectors} extended the fundamental and well-known theorem in spectral graph theory relating cardinality of zero eigenvalue of the Laplacian of a graph to the number of connected components to the uniform hypergraphs. Specifically, they proved that the algebraic multiplicity of zero eigenvalue of a symmetric Laplacian tensor is equal to the sum of the number of even-bipartite connected components and the number of connected components excluding the number of singletons in the given hypergraph. Such insights can not be revealed from the clique reduction methods and its variants \citep{paper:agarwal_2006}. In the machine learning community, tensor representation of hypergraphs has not gained much attention, except for a few works \citep{shashua2006multi,benson2019three}. In this work, we utilize the tensor representation of hypergraphs for detecting densely connected components.    
\subsection{Our Contributions \& Outline}
The preliminaries of hypergraph reduction to graphs and tensor representation are covered in Section \ref{sec:prelims}. We make the following contributions in this work:
\begin{itemize}
\setlength\itemsep{-0.5 mm}
    \item In Section \ref{sec:hyred_tencon}, we utilize the tensor representation of hypergraphs and prove that hypergraph reduction is a special case of tensor contraction. We propose the ratio-cut and normalized cut objective functions in Section \ref{sec:sub_rcuthy}, which are capable of distinguishing the multiple ways of cutting a hyperedge.
    \item In Section \ref{sec:part_algo}, we prove that the solution to the minimization of relaxed ratio-cut problem can be obtained from the eigenvector corresponding to the minimum positive eigenvalue of the Laplacian tensor. We also derive an upper bound on the minimum positive eigenvalue in terms of the conductance of the hypergraph, which is a significant improvement over the existing bounds. 
    \item In Section \ref{sec:teneig_com}, we exploit the structure of the Laplacian tensor to reduce the objective function computation time from $O(n^k)$ to $O(m)$, where $n,m,k$ are the number of nodes, hyperedges, and cardinality of hyperedges, respectively. In Section \ref{sec:exp}, we demonstrate the efficacy of the proposed algorithm on synthetic hypergraphs generated by stochastic block model. We also report $n/8$ times improvement of ratio-cut over the conventional spectral partitioning for \textit{cockroach graph}.  
\end{itemize}
The proofs of all theoretical claims and numerical details of examples are given in Appendix. 

%% file: sections/prelims.tex
\section{Preliminaries}
\label{sec:prelims}
In this section, we briefly discuss the prevalent approach of representing hypergraphs and their partitioning. A hypergraph $G$ is defined as a pair of $G = (V, E)$, where $V = \{v_1, v_2, \ldots, v_n \}$ is the set of entities called vertices or nodes and $E = \{e_1, e_2, \ldots, e_m\}$ is a set  of non-empty subsets of $V$ referred to as hyperedges. 
The strength of interaction among nodes in the same hyperedge is quantified by the positive weight represented by $w_e = \{w_{e_1}, w_{e_2}, \ldots,  w_{e_m} \}$. 

The vertex-edge incidence  matrix  is  denoted  by $\mathbf{H}$  and has  the  dimension $|V| \times |E|$.   The  entry $h(i,j)$ is defined to be $1$ if $v_i \in e_j$ and $0$ otherwise.
The degree of node $v_i$ is defined by $d(v_i) = \sum_{e_j \in E} w_{e_j} h(i, j)$. We can also define two diagonal matrices, $\mathbf{W}$, $\mathbf{D}$, with the dimension of $m \times m$, $n \times n$, containing the hyperedge weights and node degrees respectively. 
Note that there is no loss of information in this form of representation of hypergraphs until this point. This implies that a unique hypergraph can be constructed for a given incidence matrix. 
\subsection{Reducing a Hypergraph to a Graph}
Now, we discuss the widely-accepted approach for hypergraph reduction in the machine learning community. The fundamental idea is to reduce a hypergraph to graph and subsequently apply standard graph-based algorithms. In this subsection, we briefly discuss the merits and demerits of these approaches and articulate the reasons for choosing the tensor based representation of hypergraphs. 
\begin{definition}
The clique expansion for hypergraph $G(V,E)$ builds a graph $G_x(V, E_x \subseteq V^2)$ by replacing each hyperedge with the corresponding clique, $E_x = \{ (v_i, v_j) : v_i, v_j \in e_l, e_l \in E \} $ \citep{paper:agarwal_2006}. The edge weight $w_x(u, v)$ is given by $w_x(u, v) = \sum_{u,v \in e_l, e_l \in E} w(e_l)$.
\end{definition}
The same could be stated in matrix form as  
\begin{align}
\mathbf{A} = \mathbf{HW}\mathbf{H}^T - \mathbf{D} \label{eq:hyred_cliq}
\end{align}
where $\mathbf{A}$ represents the adjacency matrix for reduced hypergraph. 
Another traditional hypergraph reduction approach is star expansion \citep{paper:zien1999multilevel}. Most of the other reduction approaches are build on these. Please see  \citep{paper:agarwal_2006} and the references therein, for more details. 

This reduction step is very convenient as we can now employ any graph learning algorithms that  scale well and come with theoretical guarantees. A natural question arises on the need for these different reduction based approaches. We believe that each of these reduction approaches preserves a few \textit{but} not all hypergraph properties in the reduction step. The preserved hypergraph property may be useful for the end task of learning on hypergraphs. For example, \citep{paper:Tarun} demonstrates the improvement in clustering by proposing a reduction approach which preserves the node degrees. 

More often, the reduction step loses vital information about hypergraphs as two different hypergraphs can reduce to the same graph. This can be seen directly from Eq. \eqref{eq:hyred_cliq} as two distinct hypergraphs having different $\mathbf{H}$ and $\mathbf{W}$ can reduce to the same adjacency matrix $\mathbf{A}$. An illustrative example of the same is presented in Appendix \ref{sec:appn_redsameGr}. 

We discuss the demerits of reduction-based approaches in the context of hypergraph partitioning in the later sections. In Section \ref{sec:hyred_tencon}, we use the tensor representation of hypergraphs, and show that reduction is just a special case of tensor contraction. 

\subsection{Tensor representation of Hypergraphs}
In this subsection, we briefly review the tensor-based representation of hypergraphs \citep{qi2017tensor,banerjee2017spectra}. A natural representation of hypergraphs is a $k$-order $n$-dimensional tensor $\mathcal{A}$, which consists of $n^{k}$ entries and is defined by:
\begin{align}
	a_{i_1 i_2\ldots i_k} = 
	\begin{cases}
   	\frac{w_{e_j}}{(k-1) !} & \text{if} \quad \{i_1, i_2, \ldots, i_k\} \in E, \quad 1 \leq i_1, \ldots, i_k \leq n  \\
	0 & \text{otherwise}
	\end{cases}
\end{align}
It should be noted that $\mathcal{A}$ is a ``\emph{super-symmetric}" tensor, i.e, $a_{i_1i_2\ldots i_k} = a_{\sigma \left(i_1i_2\ldots i_k\right)} $, where $\sigma \left(i_1, i_2, \ldots i_k\right)$ denotes any permutation of the elements in the set $\{i_1, i_2, \ldots ,i_k \} $. The order or mode of the tensor refers to the hyperedge cardinality, which is $k$ for $\mathcal{A}$.  
The degree of all the vertices can be represented by $k$-order $n$-dimensional diagonal tensor $\mathcal{D}$. The Laplacian tensor $\mathcal{L}$ is defined as follows: 
\begin{align}
	\mathcal{L} = \mathcal{D} - \mathcal{A} \label{eq:lap_t_den}
 \end{align}
An example demonstrating the tensor representation of a 4-uniform hypergraph is presented in Appendix \ref{app:def_hyp_ex}. The normalized Laplacian tensor, denoted by $\mathscr{L}$ can also be defined in a similar manner:
\begin{align}
\ell_{i_1 i_2\ldots i_k} = 
\begin{cases}
-	\frac{w_{e_j}}{(k-1) !} \prod_{i_j=1}^{k} \frac{1}{\sqrt[k]{d_{i_j}}} & \text{if} \quad \{i_1, i_2, \ldots, i_k\} \in E \\
1 & \text{if $i_1 = i_2  \ldots = i_k = i $}, \qquad i = \{1,2,\ldots,n\} \\ 
0 & \text{otherwise}
\end{cases}
\label{eq:Lt_ent_norm}
\end{align}
For the sake of completeness, we define the tensor eigenvalue decomposition as: 
\begin{align}
	\mathcal{L} \mathbf{x}^{k-1} = \lambda \mathbf{x}, \qquad \text{such that} \quad \mathbf{x}^T \mathbf{x} = 1 \label{eq:ten_eig}
\end{align}
where $(\lambda, \mathbf{x}) \in (\mathbb{R}, \mathbb{R}^n \backslash \{0\}^n)$ is called the Z-eigenpair and $\mathcal{L} \mathbf{x}^{k-1} \in \mathbb{R}^n$, whose $i^{th}$ component  is: 
\begin{align}
	\left[ \mathcal{L} \mathbf{x}^{k-1} \right]_i = \sum_{i_k = 1}^{n} \ldots \sum_{i_3 = 1}^{n} \sum_{i_2 = 1}^{n} l_{ii_2i_3\ldots i_k} x_{i_2} x_{i_3} \ldots x_{i_k}  
\end{align}
In the next section, we discuss the merits of choosing tensor representation of hypergraphs over using reductions to graphs. 

%% file: sections/hy_contr.tex
\subsection{Hypergraph Reduction using Tensor Contraction}
\label{sec:hyred_tencon}

\textit{Agarwal et al}. \citep{paper:agarwal_2006} 
proposes to unify various hypergraph reduction methods like star expansion \citep{paper:zien1999multilevel} , Bolla's Laplacian \citep{bolla1993spectra}, Rodriguez’s Laplacian \citep{rodriguez2003laplacian}, Zhou's Normalized Laplacian \citep{zhou2005beyond} using clique expansion idea. In this section, we show clique expansion, and hence the other existing hypergraph reduction methods, are just a special case of tensor contraction. We define a contraction as an operation on the tensor, which reduces its mode or order. The following lemma is crucial for the proof. 
\begin{lemma}
\label{lem:zeig_zero}
One of the Z-eigenpair of $\mathcal{L}$ is $(0, \mathbf{v})$, where $\mathbf{v} = \left( \frac{1}{\sqrt{n}}, \frac{1}{\sqrt{n}}, \ldots, \frac{1}{\sqrt{n}}\right) \in \mathbb{R}^n$.
\end{lemma}
\begin{theorem}
\label{thm:contr_Tarun}
Let  $\mathbf{L_T}$ be the graph Laplacian corresponding to the clique expansion of a k-uniform hypergraph such that the node degrees are preserved \citep{paper:Tarun}.  $\mathbf{L_T}$ can be obtained from the hypergraph tensor Laplacian using $\mathbf{L_T} =  \mathcal{L} \mathbf{1}_n^{k-2}$, 
where $\mathbf{1}_n$ is a column vector of dimension $n$ with all entries as unity and $\mathcal{L} \mathbf{1}_n^{k-2} \in \mathbb{R}^{n \times n}$ with its elements defined by 
\begin{align}
\left[ \mathcal{L} \mathbf{1}^{k-2} \right]_{ij} = \sum_{i_k = 1}^{n} \ldots \sum_{i_4 = 1}^{n}  \sum_{i_3 = 1}^{n}  l_{ij i_3\ldots i_k} 
\end{align}
\end{theorem}
\begin{corollary}
\label{corr:contr_cliq}
The Laplacian corresponding to clique expansion of a k-uniform hypergraph, denoted by $\mathbf{L_c}$ can be derived from the hypergraph tensor Laplacian using $\mathbf{L_c} = (k-1) \mathcal{L} \mathbf{1}_n^{k-2} = (k-1) \mathbf{L}_T $.
\end{corollary}
This can also be interpreted as $(k-2)^{th}$ order contraction of original $k$-order tensor. 
In this section, the generalized nature of tensor Laplacian was demonstrated. This was done by deriving the mapping for $(k-2)$ order contraction of original hypergraph tensor Laplacian of order $k$. This operation reduces a $k^{th}$ order tensor to a matrix, which can be viewed as a $2^{nd}$ order tensor. 

Having established the generality of thee tensor representation, we revisit the original problem of hypergraph partitioning in the next section.  

%% file: sections/prop_algo.tex
\section{Partitioning of Hypergraphs}
\label{sec:hy_part}
We start this section with brief review of spectral graph theory for partitioning of graphs \citep{von2007tutorial} and further propose these ideas for hypergraphs. 
\subsection{Partitioning of Graphs}
Let the $p$ partitions of vertex set $V$ be denoted by sets $C_1, C_2, \ldots, C_p$ such that
\begin{align}
     C_i \neq \emptyset,\quad C_i \subset V , \quad \cup_{i=1}^{p}C_i = V, \quad C_i \cap C_j =  \emptyset, \quad  \forall i, j \in [p], \quad \text{and} \quad i \neq j \label{eq:part_defn}
\end{align}
The two most commonly used objectives of graph partitioning are Ratio cut \citep{hagen1992new} and Normalized cut \citep{shi2000normalized}:
\begin{align}
    	\text{Ratio Cut}(C_1, C_2, \ldots, C_p) =&  \sum_{i = 1}^{p} \frac{\text{cut} (C_i, \bar{C_i})}{2|C_i|},   \quad \text{where} \quad  \text{cut} (C_i, \bar{C_i}) = \sum_{r \in C_i, s \in \bar{C_i}} w_{rs} \label{eq:graph_Rcut} \\
    	\text{Normalized Cut}(C_1, C_2, \ldots, C_p) &=  \sum_{i = 1}^{p} \frac{\text{cut} (C_i, \bar{C_i})}{2 \text{vol}(C_i)}, \quad  \text{where} \quad \text{vol}(C_i) = \sum_{r \in C_i} d_r \label{eq:graph_Ncut}
\end{align}
where $w_{rs}$ denotes the weight of edge between nodes $r$ and $s$, and $d_r$ denotes the degree of $r^{th}$ node. It is well known that the solution to the relaxed version of minimizing the ratio cut and normalized cut can be obtained from the Fiedler vector of unnormalized and normalized Laplacians, respectively. 
The approximation made in the relaxation step is theoretically analyzed by \citep{chung2007four}. Several extension of this work can be seen in \citep{buhler2009spectral,lee2014multiway}. 


\subsection{Ratio-Cut and Normalized-Cut for Hypergraphs}
\label{sec:sub_rcuthy}
We start the discussion with a formal description of the problem. Let $C_1, C_2, \ldots, C_p$ be the $p$ partitions as defined in Eq. \eqref{eq:part_defn}. For a given hypergraph $G(V,E,W_e)$, we intend to remove a subset of hyperedges $\partial E \subseteq E$, such that $G \setminus \partial E$ produces at least $p$ disjoint partitions \citep{paper:kway_hypergraph2018,chekuri2015note}.  The hyperedge boundary $\partial E $ can be defined as:
\begin{align}
    \partial E = \{e_j \in E: e_j \cap C_i \neq \varnothing , e_j \cap \bar{C_i} \neq \varnothing \} 
\end{align}
The next step is to define the objective function to be minimized for obtaining optimal partitions. The measures described in Eq. \eqref{eq:graph_Rcut} and Eq. \eqref{eq:graph_Ncut} for graphs are not well-suited for hypergraphs. We propose the generalization of ratio-cut and normalized-cut for hypergraphs. 

\begin{definition}
The cut cost for the partition $C_i$ denoted by $w_h(C_i)$ and the total cut cost denoted by $w_{h,t}(V)$ for all the partitions is defined as:
\begin{align}
	w_h(C_i) =  \sum_{ e_j \in \partial E} |C_i \cap e_j| w_{e_j}, \qquad  w_{h,t}(V) = \frac{1}{k} \sum_{i = 1}^{p} w_h(C_i)  \label{eq:cost_1part}  
\end{align}
\end{definition}
The cut cost for a partition and total cut cost defined in Eq.  \eqref{eq:cost_1part}  reduces to numerator term in Eq. \eqref{eq:graph_Rcut} and Eq. \eqref{eq:graph_Ncut} for $k=2$ because the term   $ |C_i \cap e_j|$ reduces to unity $\forall e_j \in \partial E$ in graphs. We further demonstrate the merits of this cut cost by the following example. 

\begin{minipage}{0.4 \textwidth}
\begin{figure}[H]
    \centering
    \input{figs/ex_3uni1.tex}
    \captionsetup{justification=centering}
    \caption{Hypergraph : $H_1$}
    \label{fig:3uni_r1}
\end{figure}
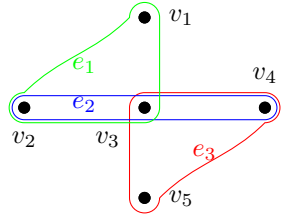
\end{minipage}
\begin{minipage}{0.5 \textwidth}
\begin{center}
\begin{figure}[H]
	\centering
    \input{figs/ex_3uni2part.tex}
    \captionsetup{justification=centering}
    \caption{Hypergraph : $H_2$}
    \label{fig:3uni_2part}

\end{figure}
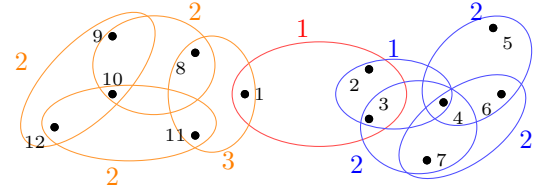
    \end{center}
\end{minipage}

\begin{example}
Consider the 3-uniform hypergraph shown in Figure \ref{fig:3uni_r1}. 
Consider the partitions obtained after removing hyperedges  $e_2$ and $e_3$. Let $C_1 = \{v_1, v_2, v_3\}, C_2 = \{4\}, C_3 = \{5\}$. The partition cost is given by 
\begin{align*}
    w_h(C_1) = 2 w_{e_2} + w_{e_3}, \quad  w_h(C_2) = w_{e_2} +  w_{e_3},  \quad w_h(C_3) = w_{e_3}, \quad w_{h,t}(V) =  w_{e_2} + w_{e_3}
\end{align*}
It should be noted that $w_{e_1}$ is not reflected in the above cut costs because hyperedge $e_1$ is not cut. It could be easily verified that this cut cost is \textit{not} equivalent to clique reduction approach. The cut costs derived for the reduced hypergraph are as follows:
\begin{align}
     w_g(C_1) = 2 (w_{e_2} + w_{e_3}), \quad  w_g(C_2) = 2(w_{e_2} +  w_{e_3}),  \quad w_g(C_3) = 2w_{e_3}, \quad w_{h,t}(V) =  2w_{e_2} + 3w_{e_3} \nonumber
\end{align}
Note that the cut costs derived from both approaches are different. On further inspection, we infer $w_g(C_i) = 2w_h(C_i)$ for $i=\{2,3\}$, which means the cut cost for partitions $C_2$ and $C_3$ in the reduced hypergraph are just a scaled version of costs involved in original hypergraph. The same relation does not hold for partition $C_1$ due to the presence of the term $|C_i \cap e_j|$ in Eq. \eqref{eq:cost_1part}. Please refer Appendix \ref{app:partcost} for the computation of these cut costs. 
\end{example}
From this illustrative example, it can be inferred that the proposed cut cost for hypergraphs defined in Eq. \eqref{eq:cost_1part} comprises of more information about the cut as compared to reduced hypergraphs. The term $|C_i \cap e_j|$  in Eq. \eqref{eq:cost_1part} will lead to a greater penalty for cutting hyperedges with more elements from $C_i$. 
A hyperedge with higher $|C_i \cap e_j|$ is likely to have more association with partition $C_i$, so the corresponding cut should be penalized more. 

Minimizing the total cut cost defined in Eq. \ref{eq:cost_1part} directly may lead to ``unbalanced'' partitions with minimum cost. To bypass such trivial and undesirable partitions, we propose the normalization.

\begin{definition}
\label{prop:rcut_ncut_def}
The Ratio-cut and Normalized-cut for $p$ partitions are defined as:
\begin{align}
    \text{Ratio-Cut}(C_1, C_2, \ldots, C_p) = \sum_{i = 1}^{p} \frac{w_h(C_i)}{k|C_i|^{k/2}} , \quad 
    \text{N-Cut}(C_1, C_2, \ldots, C_p) =   \sum_{i = 1}^{p} \frac{w_h(C_i)}{k(\text{vol}(C_i))^{k/2}} \label{eq:min_rcut} 
\end{align}
\end{definition}
where $w_h(C_i)$ is defined in Eq.\eqref{eq:cost_1part}. The above term for ratio-cut and normalized-cut simplifies to Eq.\eqref{eq:graph_Rcut} and Eq.\eqref{eq:graph_Ncut} respectively for $k = 2$. We prefer normalization of exponential factor in the denominator to bypass the partitions with singletons or less number of nodes.
\citep{ghoshdastidar2015provable, ghoshdastidar2015spectral} define normalized  associativity of a hypergraph in a similar manner but utilize the approach of hypergraph reduction for modelling purposes. 

\subsection{Hypergraph Partitioning Algorithm}
\label{sec:part_algo}
The partitions $C_1, C_2, \ldots, C_p$ can be derived by minimizing the ratio-cut or normalized cut. For further discussion, we focus on the minimization of ratio-cut, and the same approach can be extended for normalized-cut, as shown later. The optimal partitions can be obtained by solving:
\begin{align}
 {(C_1, C_2, \ldots, C_p)} =   \argmin_{(C_1, C_2, \ldots, C_p)}  \frac{1}{k}\sum_{i = 1}^{p} \frac{w_h(C_i)}{|C_i|^{k/2}} 
    \label{eq:opti_hy_ratio}
\end{align}
Unfortunately, the above problem is NP-hard. Inspired from spectral graph theory, we propose to solve a relaxed version of the optimization problem mentioned above. 
\begin{theorem}
\label{thm:Rcut_teig}
The minimization of ratio-cut in Eq. \eqref{eq:min_rcut} can be equivalently expressed as
\begin{align}
   \min \left(\sum_{i=1}^{p}	\mathcal{L} \mathbf{f}_i^{k} \right)= \min \left( \sum_{i=1}^{p} |C_i \cap e_j|  \frac{w_{e_j} }{|C_i|^{k/2}}\right), \qquad f_{i,j}= 
\begin{cases}
\frac{1}{\sqrt{|C_j|}} &\quad v_i \in C_j \\
0 & \text{otherwise}
\end{cases} \label{eq:Rcut_hy_disc}
\end{align}
where we define $p$ indicator vectors $\mathbf{f}_i$ and its $j^{th}$ element, denoted by $f_{i,j}$ indicates if the vertex $v_i$ belongs to cluster $C_j$. The solution to the above problem after relaxing $\mathbf{f}_i \in \mathbb{R}^n$ rather than an indicator vector can be derived from the eigenvector corresponding to the minimum positive eigenvalue  stated in Eq. \eqref{eq:ten_eig}. 
\end{theorem}
Hence, we focus on the computation of objective function $ \mathcal{L} \mathbf{x}^{k}$ for any vector $\mathbf{x}\in \mathbb{R}^{n}$. 
\begin{theorem}
\label{thm:Lxk_comp}
The objective for the tensor Laplacian of a hypergraph can be simplified using 
{\small 
\begin{align}
	\mathcal{L} \mathbf{x}^{k} &= \sum_{i_1,i_2,\ldots, i_k = 1}^{n} l_{i_1i_2\ldots i_k} x_{i_1} x_{i_2} \ldots x_{i_k} \nonumber \\
	&= \sum_{e_j \in E} w_{e_j} \left( \sum_{i_t \in e_j } x^k_{i_t} - k \prod_{i_t \in e_j} x_{i_t} \right) = \sum_{e_j \in E} w_{e_j} k \left( \underset{ i_t \in e_j }{\text{A.M}\left(x^k_{i_t} \right)} -  \underset{ i_t \in e_j }{\text{G.M}\left(|x_{i_t}|^k \right)} (-1)^{n_s} \right) 
	\label{eq:am_gm_main}
\end{align}}
where $n_s = |\{ i_j : x_{i_j} < 0 \}|$, A.M and G.M stand for arithmetic and geometric means respectively. 
\end{theorem}
The above objective function can be viewed as generalization of the graph, as for any edge $\{a,b\}$, the objective function $(x_a - x_b)^2 = x_a^2 + x_b^2 - 2 x_a x_b$.  
We continue the discussion on partitioning  with the following example. 
\begin{example}
\label{ex:3uni_2part}
Consider the 3-uniform hypergraph shown in Figure \ref{fig:3uni_2part}. The colored number indicates the hyperedge weight. It is clear that the optimal partitions are $A_1 = \{2, 3, 4, 5, 6, 7\}$ and $\bar{A}_1$. The Fiedler eigenvector for this hypergraph is 
\begin{align*}
    \mathbf{f^{\star}} = [
    0.33 \quad 0.16 \quad 0.17 \quad 0.13 \quad -0.05 \quad 0.05 \quad 0.12 \quad 0.39 \quad 0.38 \quad 0.43 \quad 0.39 \quad 0.38 ]
\end{align*}
A standard approach in spectral graph theory is to use the sign of the elements in the Fiedler vector for partitioning \citep{von2007tutorial}. For example, $C_1 = \{ i | \mathbf{f}^{\star}(i) < 0, i \in [n] \}$.
Hence, the partitions are $C_1 = \{5\}$ and $C_2 = V \setminus C_1$, which is clearly not the optimal. 
\end{example}
From the above example, it is clear that the traditional approach of partitioning does not yield desired partitions for hypergraphs. This is primarily because the eigenvectors of the Laplacian tensor of a hypergraph can not be interpreted in the same way as the eigenvectors of the Laplacian matrix of a graph. 

To understand the implication of minimum ratio-cut associated with minimum positive $\lambda^{\star} $, we analyze the computation of Laplacian objective function using the Fiedler vector:
\begin{align}
	l_{e_j} (\mathbf{f^{\star}})= w_{e_j}  \left( \sum_{i_k \in e_j } f^k_{i_k} - k \prod_{i_k \in e_j} f_{i_k} \right), \qquad 
	\lambda^{\star}  = \sum_{e_j \in E} l_{e_j} (\mathbf{f^{\star}}) \label{eq:L_edge} 
\end{align}
where $l_{e_j}(\mathbf{f^{\star}})$ denotes the ``score" for hyperedge $e_j$ computed for the eigenvector $\mathbf{f^{\star}}$. With a slight abuse of terminology, we argue that a higher value of this score indicates the corresponding hyperedges are ``close" to separator boundary  $\partial E $. The measure of closeness between two nodes is quantified by the minimum number of hyperedges to be traversed for reaching one node to another. 

This can be validated easily by careful inspection of hyperedge score $l_{e_j} (\mathbf{f})$, when the vector $\mathbf{f}$ is treated as the cluster indicator variable shown in Eq. \eqref{eq:Rcut_hy_disc}. The hyperedge score will be non-zero only for the hyperedges on the separating boundary for such ideal choice of $\mathbf{f}$. The same can be also interpreted as the score being zero $\forall e_j \in \{E \setminus \partial E\}$. We carry forward the same intuition and prefer to cut the hyperedges with a ``higher" score. 

The score may not be exactly zero for any hyperedge if the Fiedler vector is used for the score computation as it is obtained for the relaxed minimization of the ratio-cut (Theorem \ref{thm:Rcut_teig}). Applying this approach on Example \ref{ex:3uni_2part}, we report a maximum score of $0.017$ for the hyperedge $\{1,2,3\}$ and hence cut it to obtain the optimal partitions. The proposed algorithm is summarized in Table \ref{tab:algo}. 
\begin{table}[h]
	\caption{Hypergraph Partitioning Algorithm }
	\label{tab:algo}
	\vspace{-1mm}
		\hrulefill
		\vspace{-3mm}
	\begin{enumerate}
	\itemsep 0.01mm
		\item Construct the tensor Laplacian and derive the Fiedler eigenpair $(\lambda^{\star} ,\mathbf{f}^{\star})$.
		\item Calculate the hyperedge score, $l_{e_j}(\mathbf{f}^{\star})$ by using Eq. \eqref{eq:L_edge}.
		\item Remove hyperedges with maximum cost until p-disjoint partitions are obtained.
		\vspace{-1mm}
	\end{enumerate}
\hrule
\end{table}

The intuition behind using the hyperedge score for deriving $\partial E $ is motivated from spectral graph theory. It is interesting to note that this novel use of hyperedge scores helps to compute a better ratio-cut for the cockroach graph presented in Section \ref{sec:exp}.

A similar analysis can be performed for the minimization of the normalized cut of hypergraphs. 
\begin{corollary}
\label{cor:normcut_teneig}
The solution to the relaxed optimization problem of minimizing normalized cut mentioned in Proposition \ref{prop:rcut_ncut_def} can be derived using the eigenvector corresponding to the minimum positive eigenvalue of the normalized Laplacian tensor defined in Eq. \eqref{eq:Lt_ent_norm}. 
\end{corollary}
We perform the theoretical analysis of the proposed algorithm and derive an interesting bound on the approximation made in normalized cuts. 

\begin{theorem}
\label{thm:lambda_bound}
The upper bound on the minimum positive eigenvalue of an even order $k$-uniform hypergraph is
\begin{align}
	\lambda_1  \leq k \phi(G), \qquad \phi(G) = \min_{C \subseteq V}\frac{\sum_{e_j \in {\partial E}} w_{e_j}}{\min \left\{ \text{vol}(C), \text{vol}(\bar{C})\right\}}, \qquad \text{vol}(C) = \sum_{i_j \in C} d_{i_j} \label{eq:prop_cheeger}
\end{align}
where $\lambda_1$ is the first positive eigenvalue satisfying Eq. \eqref{eq:ten_eig} for normalized tensor Laplacian $\mathscr{L}$ and $\phi(G)$ refers to the conductance of hypergraph.
\end{theorem}
The above inequality helps to analyze the order of approximation involved in relaxing the N-min cut problem by deriving the solution through tensor EVD. The tightness of the bound indicates the goodness of the approximation. 
Several other attempts have been made to derive such approximation bounds for hypergraphs. For example, 
\citep{chen2017fiedler} utilizes a different Laplacian tensor and the following hyperedge score to derive similar bound on $\lambda_1$ of a different tensor:
\begin{align}
    l_{e_j} (\mathbf{x}) = \sum_{i_k \in e_j} (x_{i_k} - \bar{x})^k, \quad \bar{x} = \frac{1}{k}\sum_{i_k \in e_j} x_{i_k}, \qquad 	\lambda_1  \leq 2^{k/2} \phi(G) \label{eq:cheeg_ten}
\end{align}
This is a weaker bound of exponential nature where as we have proposed tighter bound of linear nature in Theorem \ref{thm:lambda_bound}.

\subsection{Computation of Tensor Eigenvectors}
\label{sec:teneig_com}
The computation of eigenvectors of real super-symmetric tensors is quite challenging and not straightforward as in the case of real symmetric matrices. This is primarily due to the non-orthogonality of tensor eigenvectors. There are several other challenging aspects, for example,  real symmetric tensors can have complex eigenpairs, unlike the case of matrices. Also, a real symmetric matrix of size $n \times n$ can have a maximum of $n$ eigenvalues, whereas a tensor can have much larger number of eigenpairs \citep{qi2005eigenvalues}. Please refer to \citep{qi2017tensor} and references therein for more details.  

Most of the existing works on computation of eigenpairs have been for tensors with special structure \citep{qi2009z} or the extreme eigenvalues such as maximum or minimum eigenvalue \citep{kolda2011shifted, hu2013finding}. As discussed in the section \ref{sec:part_algo}, only the Fiedler vector is required for partitioning a given hypergraph. As the Fiedler vector is not one of the extreme eigenvectors, the above methods are not helpful for our case.   

Recently, \citep{chen2016computing} proposed an algorithm to compute all the eigenvalues of a tensor using homotopy methods. They pose the problem as finding the roots of a vector of high order polynomials generated from $P(\mathbf{y}) = \mathcal{L}\mathbf{x}^{k-1} - \lambda \mathbf{x} = \mathbf{0}$, where $\mathbf{y} = [\mathbf{x} \quad \lambda] \in \mathbb{R}^{n+1}$. As it is tough to compute the zeros of  $P( \mathbf{y})$ directly, the core idea of linear homotopy methods is to construct a vector function $H(\mathbf{y},t) = (1-t) Q(\mathbf{y}) + t P(\mathbf{y})$, where $t \in [0,1]$ and $Q(\mathbf{y})$ is a suitable vector polynomial whose roots can be computed easily.  The next step is to slowly iterate from the solution of $H(\mathbf{y},t=0) =  Q(\mathbf{y}) = \mathbf{0}$ to $H(\mathbf{y},t=1) =  P(\mathbf{y}) = \mathbf{0}$. Despite the novel formulation, this approach is forced to compute all the complex eigenpairs even if we are interested in real eigenpairs only.  

We prefer to use the approach by \citep{cui2014all}, which computes all the real eigenvalues sequentially from maximum to minimum by using Jacobian semidefinite relaxations in polynomial optimization. They formulate the following problem to compute $\lambda_{i+1}$ assuming $\lambda_i$ is known:
\begin{align}
    \max & \quad f(\mathbf{x}) = \mathcal{L}\mathbf{x}^{k} \nonumber \\
    \text{such that} & \quad f(\mathbf{x}) \leq \lambda_k  - \delta \quad \& \quad  h_r(\mathbf{x}) = 0, \quad (r = 1, \ldots, 2n-2)  \label{eq:eigcomp}
\end{align}
where $0 < \delta < \lambda_i - \lambda_{i+1}$ and $h_r(\mathbf{x})$ is defined as:
\begin{align}
    h_r(\mathbf{x}) = \sum_{i+j = r+2} \frac{\partial  f(\mathbf{x})}{\partial x_i}  \frac{\partial g(\mathbf{x})}{\partial x_j}  -  \frac{\partial f(\mathbf{x}) }{\partial x_j}  \frac{\partial g(\mathbf{x})}{\partial x_i} 
\end{align}
where $g(\mathbf{x}) = x_1^2 + x_2^2 + \ldots + x_n^2 - 1$ is a normalization constraint. They further utilize Lasserre's hierarchy of semidefinite relaxations \citep{lasserre2001global} to solve the above problem. 

The computation of the objective function  $f(\mathbf{x})$ and the constraints $ h_r(\mathbf{x}) $ is expensive and takes $O(n^k)$ for general tensors. Using Theorem \ref{thm:Lxk_comp}, the objective function can be computed in linear time $O(m)$ for Laplacian tensors. The constraint can also be simplified using:
\begin{align}
   \frac{\partial  f(\mathbf{x})}{\partial x_i}  = \sum_{e_p \in E_i}k w_{e_p} \left( x^{k-1}_{i} - k \prod_{t \in \{e_p \setminus i\}} x_{t} \right)
\end{align}
where $E_i = \{ e_q | i \cap e_q \neq \emptyset, e_q \in E\}$. This approach is very helpful as all the eigenvalues need not be computed for the Fiedler eigenvalue. Hence, the Fiedler vector can be computed easily by using Theorem \ref{thm:Lxk_comp} and the above equation.  
\subsection{Related works}
As stated earlier, most of the existing methods utilize hypergraph reductions either implicitly \citep{paper:agarwal_2006,zhou2007learning} or explicitly. For example, \citep{ghoshdastidar2014consistency} utilize the tensor-based representation of hypergraphs but construct a matrix by concatenating the slices of the tensor. Further, they apply the standard spectral partitioning algorithm on the covariance of that matrix. These variants of hypergraph reduction differ in the method of expanding a hyperedge and produce graphs with different edge weights. The Laplacian objective function (Eq.\eqref{eq:am_gm_main}) of any graph is second-order polynomial, which captures weighted interaction among two nodes. A second-order polynomial is insufficient for capturing super-dyadic interaction among multiple nodes ($\geq 3$) of a hyperedge. Also, note that multiple hypergraphs may reduce to the same graph. 

\citep{hein2013total} discuss the incapability of reduction methods in preserving the hyperedge cuts for general hypergraphs. We utilize the Laplacian tensor (Eq. \eqref{eq:am_gm_main}) to penalize these multiple cuts differently. Few other recent works try to capture these multiple ways of splitting nodes. For example, \citep{li2017inhomogeneous} proposes non-uniform clique expansion and provides quadratic approximation under submodularity constraints of the inhomogeneous cost function. \citep{li2018submodular} extends the notion of p-Laplacian from graphs to hypergraphs by introducing the following hyperedge score:
\begin{align}
l_{e_j}(\mathbf{x}) = \max_{i_k,i^{'}_k \in e_j} |x_{i_k} - x_{i^{'}_k}|^p \label{eq:maxdiff}
\end{align}
Ideally, any definition of hyperedge score should capture the non-uniformity among the nodes in a hyperedge, but the above equation fails to capture the variation perfectly. For example, consider two hyperedges with cardinality 4 and node labels assigned as $\{0,1,1,2\}$ and $\{0,1,2,2\}$. Equation \eqref{eq:maxdiff} computes the maximum difference and hence will not differentiate among these two hyperedges but the AM-GM difference (Eq. \eqref{eq:L_edge}) will capture the variance among all the nodes of the hyperedge. \citep{zhang2017re, chan2018spectral, chan2020generalizing} consider a similar formulation of the hyperedge score function.

%% file: figs/ex_3uni1.tex
\begin{tikzpicture}[scale = 0.8]
    \node (v3) at (0,0) {};
    \node (v2) at (-2,0) {};
    \node (v4) at (2,0) {};
    \node (v1) at (0,1.5) {};
    \node (v5) at (0,-1.5) {};
    
    \node (e1) at (-1,0.7) {\textcolor{green}{$e_1$}};
    \node (e1) at (-1,0.05) {\textcolor{blue}{$e_2$}};
    \node (e1) at (1,-0.75) {\textcolor{red}{$e_3$}};
    
    \begin{scope}[fill opacity=0.8]
    
    \draw[red!100] ($(v5)+(0.25,0)$)
        to[out=45,in=225] ($(v4) + (0,-0.25)$) 
        to[out=0,in=270] ($(v4) + (0.25,0)$) 
        to[out=90,in=0] ($(v4) + (0,0.25)$)
        to[out=180,in=0] ($(v3) + (0,0.25)$) 
        to[out=180,in=90] ($(v3) + (-0.25,0)$)
        to[out=270,in=90] ($(v5) + (-0.25,0)$) 
        to[out=270,in=180] ($(v5) + (0,-0.25)$) 
        to[out=0,in=270] ($(v5) + (0.25,0)$) ; 
    
    \draw[blue!100] ($(v2)+(0,-0.2)$) 
        to[out=0,in=180] ($(v4) + (0,-0.2)$) 
       to[out=0,in=270] ($(v4) + (0.2,0)$)
        to[out=90,in=0] ($(v4) + (0,0.2)$)
        to[out=180,in=0] ($(v2) + (0,0.2)$)
        to[out=180,in=90] ($(v2) + (-0.2,0)$)
        to[out=270,in=180] ($(v2) + (0,-0.2)$);
        
    \draw[green!100] ($(v2)+(0,-0.25)$) 
         to[out=0,in=180] ($(v3) + (0,-0.25)$) 
         to[out=0,in=270] ($(v3) + (0.25,0)$)
         to[out=90,in=270] ($(v1) + (0.25,0)$)
         to[out=90,in=0] ($(v1) + (0,0.25)$)
         to[out=180,in=90] ($(v1) + (-0.25,0)$)
         to[out=225,in=45] ($(v2) + (0,0.25)$)
         to[out=180,in=90] ($(v2) + (-0.25,0)$)
         to[out=270,in=180] ($(v2) + (0,-0.25)$);
    \end{scope}
    
    \foreach \v in {1,2,...,5} {
        \fill (v\v) circle (0.1);
    }
    
    \fill (v1) circle (0.1) node [right =0.2cm] {$v_1$};
    \fill (v2) circle (0.1) node [below =0.2cm ] {$v_2$};
    \fill (v3) circle (0.1) node [below left= 0.2cm] {$v_3$};
    \fill (v4) circle (0.1) node [above= 0.2cm] {$v_4$};
    \fill (v5) circle (0.1) node [right = 0.2cm] {$v_5$};
    
\end{tikzpicture}

%% file: figs/ex_3uni2part.tex
\begin{tikzpicture}[scale = 1.1]
\def\i{1.4}
\def\bsize{0.05}

    \node (v1) at (-0.4 ,0 ) {};
    \node (v2) at (1.1 ,0.3 ) {};
    \node (v3) at (1.1 ,-0.3 ) {};
    \node (v4) at (2 ,-0.1 ) {};
    \node (v5) at (2.6 ,0.8 ) {};
    \node (v6) at (2.7 ,0 ) {};
    \node (v7) at (1.8 ,-0.8 ) {};
    \node (v8) at (-1 ,0.5 ) {};
    \node (v9) at (-2 ,0.7 ) {};
    \node (v10) at (-2 ,0 ) {};
    \node (v11) at (-1 ,-0.5 ) {};
    \node (v12) at (-2.7 ,-0.4 ) {};
    
    \draw [red!70] (0.5 ,0 ) ellipse (30pt and 18pt); 
     
    \draw [blue!70] [rotate=45](1.9 ,-1.5 ) ellipse (22pt and 15pt); 

    \draw[blue!70][rotate=35](1.6 ,-1.6 ) ellipse (25pt and 13pt); 
    
    
    \draw [blue!70] (1.4 ,0 ) ellipse (20pt and 12pt); 
    \draw [blue!70](1.7 ,-0.4 ) ellipse (20pt and 16pt); 
    
    \draw [orange!80](-0.8 ,0 ) ellipse (15pt and 20pt); 
     \draw [orange!80] (-1.8 ,-0.35 ) ellipse (30pt and 13pt); 
    \draw [orange!80] (-1.5 ,0.35 ) ellipse (21pt and 17pt); 
    \draw [orange!80][rotate=45] (-1.5 ,1.75 ) ellipse (30pt and 13pt);

    \foreach \v in {1,2,...,12} {
        \fill (v\v) circle (\bsize);
    }

    \fill (v1) circle (\bsize) node [right] {\scriptsize{$1$}};
    \fill (v2) circle (\bsize) node [below left] {\scriptsize{$2$}};
    \fill (v3) circle (\bsize) node [above right] {\scriptsize{$3$}};
    \fill (v4) circle ( \bsize) node [below right] {\scriptsize{$4$}};
    \fill (v5) circle ( \bsize) node [below right] {\scriptsize{$5$}};
    \fill (v6) circle ( \bsize) node [below left] {\scriptsize{$6$}};
    \fill (v7) circle ( \bsize) node [right] {\scriptsize{$7$}};
    \fill (v8) circle ( \bsize) node [below left] {\scriptsize{$8$}};
    \fill (v9) circle ( \bsize) node [left] {\scriptsize{$9$}};
    \fill (v10) circle ( \bsize) node [above] {\scriptsize{${10}$}};
    \fill (v11) circle ( \bsize) node [left] {\scriptsize{${11}$}};
    \fill (v12) circle ( \bsize) node [below left] {\scriptsize{${12}$}};

   \node at (0.3,0.8) {$\textcolor{red}{\footnotesize{1}}$};
    \node at (1.4,0.55) {$\textcolor{blue}{\footnotesize{1}}$};
    \node at (2.4,1) {$\textcolor{blue}{2}$};
    \node at (3,-0.55) {$\textcolor{blue}{2}$};
    \node at (0.95,-0.85) {$\textcolor{blue}{2}$};
    \node at (-0.6,-0.8) {$\textcolor{orange}{3}$};
    \node at (-1,1) {$\textcolor{orange}{2}$};
    \node at (-3.1,0.4) {$\textcolor{orange}{2}$};
    \node at (-2,-1) {$\textcolor{orange}{2}$};

\end{tikzpicture}

%% file: sections/exp.tex
\section{Experiments}
\label{sec:exp}
The proposed algorithm is examined on well-known and synthetic hypergraphs. The numerical details of Fiedler vector and hyperedge scores are presented in Appendix \ref{sec:app_num}. 

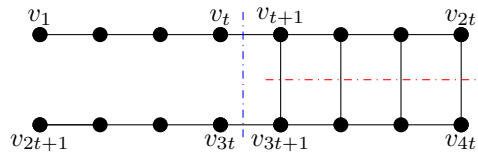
\begin{figure}[H]
    \centering
    \input{figs/cockroach.tex}
    \captionsetup{justification=centering}
    \caption{Cockroach Graph}
    \label{fig:roach}    
\end{figure}

\begin{example}
\label{ex:roach}
Consider the cockroach graph shown in Figure \ref{fig:roach} and taken from \citep{von2007tutorial}.
\end{example}

The traditional spectral partitioning makes the red cut shown in the graph and the partition is $A_1 = \{v_1, \ldots, v_{2t}\}$ and the ratio-cut$(A_1, \bar{A_1}) = \frac{t}{2t} + \frac{t}{2t} = 1$. We utilize the edge scores as suggested in the proposed algorithm and report that the edges $\{v_t, v_{t+1}\}$ and $\{v_{3t}, v_{3t+1}\}$ have maximum scores. On cutting these edges, the obtained partition is  $B_1 = \{v_1, v_2, \ldots, v_t, v_{2t+1}, \ldots, v_{3t}\}$ and hence the ratio-cut$(B_1, \bar{B_1}) = \frac{2}{2t} + \frac{2}{2t} = \frac{2}{t}$. Therefore, the solution obtained by proposed algorithm is $t/2$ times better than the traditional approach. We made this observation for $t = \{3,4,\ldots,20\}$. 

\begin{figure}[H]
\centering
\includegraphics[width=.6\textwidth]{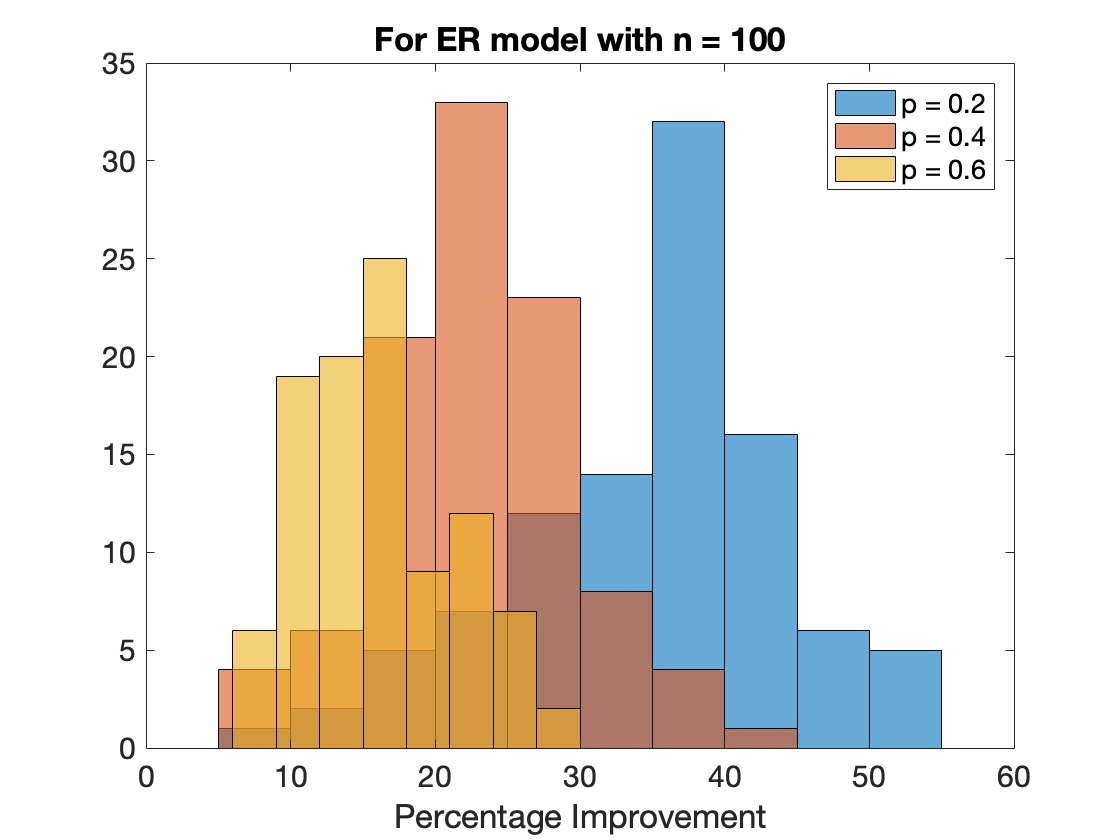}
\caption{Histogram plot for percentage improvement by the proposed method for graphs generated by the ER model for different values of $p$. It shows that the proposed algorithm performs better for all the generated graphs.}
\label{fig:hist_ER}
\end{figure}

\begin{example}
In this example, we consider different types of synthetic graphs and compare the ratio-cut values computed by the existing and proposed methods. 
\end{example}
We begin with the study on random graphs generated from the Erd\H{o}s-Rényi model denoted by $G(n,p)$ where $n$ is the number of nodes and  $p$ is the probability of an edge between any two nodes.  We compare the ratio-cut values for $100$ different graphs for $n = 100$ and for each value of $p = \{0.2,0.4,0.6\}$. We observe that the ratio-cut value by our proposed algorithm is always less than the ratio-cut obtained by sign-based Fiedler vector partitioning. Hence we define the following metric, termed as percentage improvement (PI) to showcase the proposed algorithm's performance: 
\begin{align}
    \text{PI} = \frac{(R_f - R_p)}{R_f} \times 100
\end{align}
where $R_f$, $R_p$ denotes the ratio-cut value by sign based Fiedler partitioning and proposed algorithm, respectively. A positive value of PI indicates the proposed algorithm has produced a better ratio-cut value and the magnitude of the value represents the extent of the improvement. Figure \ref{fig:hist_ER} shows the result as a histogram for different values of $p = \{0.2,0.4,0.6\}$. It can be seen that the proposed algorithm performs better than the sign based Fiedler partitioning in all the cases. 

We perform the similar analysis for another graph generation model, referred to as the stochastic block model (SBM). This model provides us the freedom to control the number of partitions, the number of nodes in each partition (denoted by $n_1, n_2$), the probability of an edge within a partition (denoted by $p$), and across the partition ($q$). Note that $p = q$ yields ER model with $n = n_1 + n_2$ as discussed previously. 

\begin{figure}[H]
\centering
\includegraphics[width=.6\textwidth]{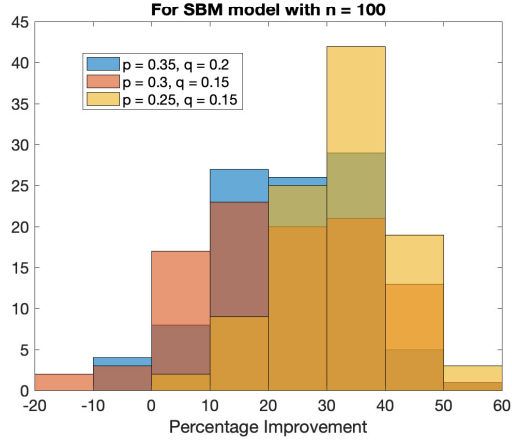}
\caption{Histogram plot for percentage improvement by the proposed method for graphs generated by the SBM for different values of intra-cluster probability (p) and inter-cluster probability (q). It confirms that the proposed algorithm performs better for most of the generated graphs as there are very few cases of negative PI.}
\label{fig:hist_sbm}
\end{figure}

We consider the graphs for multiple combinations of probabilities $p$, $q$ and $2$ partitions with $n_1 = n_2 = 50$. It should be noted that we consider the SBM with assortative community structure, which implies $p > q$. We generate 100 random graphs for each of these settings and compare the ratio-cut values. A histogram plot summarizing the results is presented in Figure \ref{fig:hist_sbm}. 

It is evident from Figure \ref{fig:hist_sbm} that the proposed algorithm produces a lower ratio-cut value for most of the graphs generated by SBM. 
We perform a similar analysis on synthetic hypergraphs generated by SBM \citep{ghoshdastidar2014consistency}. We generate $100$ random $4$-uniform hypergraphs with $2$ partitions, $60$ nodes, and relatively small values of intra-cluster probability ($p$) and inter-cluster probability ($q$) as compared to the case of graphs. This is primarily because the number of possible hyperedges for a $4-$uniform hypergraph is $\binom{n}{4}$, which is much larger as compared to the case of graphs $\left(\binom{n}{2}\right)$. 

\begin{figure}[H]
\centering
\includegraphics[width=.6\textwidth]{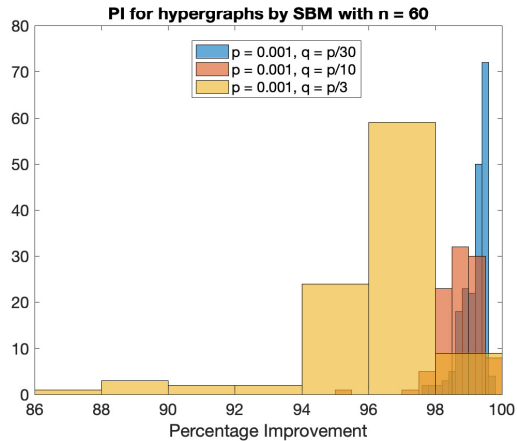}
\caption{Histogram plot for percentage improvement by the proposed method for hypergraphs generated by the SBM for different values of $q$. It shows that the proposed algorithm performs significantly better as compared to sign-based partitioning for all generated hypergraphs.}
\label{fig:hist_hySBM}
\end{figure}

The proposed algorithm is compared against the conventional sign-based partitioning using the Fiedler vector computed from the Laplacian tensor of the hypergraph. A histogram plot summarizing the results is shown in Figure \ref{fig:hist_hySBM}.

It can be observed that the proposed algorithm has improved the ratio-cut value (defined in Eq. \eqref{eq:min_rcut}) significantly as compared to the traditional sign-based partitioning. This is primarily because cutting a few hyperedges does not necessarily produce only two components, unlike the case of graphs. For example, if we cut a hyperedge having $3$ nodes in a hypergraph with one hyperedge only, we get $3$ disconnected components, and there is no possibility of obtaining two connected components. Hence, we may get $3$ connected components, even if we desired only $2$ connected components. This is a unique property of hypergraphs. 

Any partitioning algorithm producing many small connected components (like singletons) will have a higher ratio-cut value.  We observe that the conventional sign-based partitioning approach using the Fiedler vector of the Laplacian tensor is more prone to producing many small connected components as compared to the results by proposed algorithms. Hence, the ratio-cut value by sign-based partitioning is significantly higher.

In this section, we examined the performance of the algorithm on various synthetic hypergraphs and observed that the proposed approach derives the partitions with lower ratio-cut values.

%% file: figs/cockroach.tex
\begin{tikzpicture}
    \node (v1) at (0,0) {};
    \node (v2) at (0.8,0) {};
    \node (v3) at (1.6,0) {};
    \node (v4) at (2.4,0) {};
    \node (v5) at (3.2,0) {};
    \node (v10) at (0,-1.2) {};
    \node (v9) at (0.8,-1.2) {};
    \node (v8) at (1.6,-1.2) {};
    \node (v7) at (2.4,-1.2) {};
    \node (v6) at (3.2,-1.2) {};
    
    \node (v11) at (4,0) {};
    \node (v12) at (4.8,0) {};
    \node (v13) at (5.6,0) {};

    \node (v14) at (4,-1.2) {};
    \node (v15) at (4.8,-1.2) {};
    \node (v16) at (5.6,-1.2) {};
    
    \node (sp1) at (3.,-0.6) {}; 
    \node (sp2) at (5.8,-0.6) {}; 
    
    \node (p1) at (2.7,0.3) {};
    \node (p2) at (2.7,-1.4) {};
    
    \begin{scope}[fill opacity=0.8]
    \draw[] ($(v1)$) to ($(v2)$);
    \draw[] ($(v2)$) to ($(v3)$);
    \draw[] ($(v3)$) to ($(v4)$);
    \draw[] ($(v4)$) to ($(v5)$);
    \draw[] ($(v5)$) to ($(v6)$);
    \draw[] ($(v6)$) to ($(v7)$);
    \draw[] ($(v7)$) to ($(v8)$);
    \draw[] ($(v8)$) to ($(v9)$);
    \draw[] ($(v9)$) to ($(v10)$);
    \draw[] ($(v9)$) to ($(v10)$);
    
    \draw[] ($(v5)$) to ($(v11)$);
    \draw[] ($(v11)$) to ($(v12)$);
    \draw[] ($(v12)$) to ($(v13)$);
    
    \draw[] ($(v6)$) to ($(v14)$);
    \draw[] ($(v14)$) to ($(v15)$);
    \draw[] ($(v15)$) to ($(v16)$);
    
    \draw[] ($(v11)$) to ($(v14)$);
    \draw[] ($(v12)$) to ($(v15)$);
    \draw[] ($(v13)$) to ($(v16)$);
    
    \draw[dash dot, color = red] ($(sp1)$) to ($(sp2)$);
    \draw[dash dot, color = blue] ($(p1)$) to ($(p2)$);
    \end{scope}

    \foreach \v in {1,2,...,16} {
        \fill (v\v) circle (0.1);
    }

    \fill (v1) circle (0.1) node [above] {$v_1$};
    \fill (v4) circle (0.1) node [above] {$v_t$};
    \fill (v5) circle (0.1) node [above] {$v_{t+1}$};
    \fill (v6) circle (0.1) node [below] {$v_{3t+1}$};
    \fill (v7) circle (0.1) node [below] {$v_{3t}$};
    \fill (v10) circle (0.1) node [below] {$v_{2t+1}$};
    \fill (v13) circle (0.1) node [above] {$v_{2t}$};
\fill (v16) circle (0.1) node [below] {$v_{4t}$};
\end{tikzpicture}

%% file: sections/appen_proof.tex
\section{Proofs}
\label{sec:appn_proof}
\subsection{Proof of Lemma \ref{lem:zeig_zero}}
\begin{proof}
Please refer \citep{banerjee2017spectra}: Theorem 3.13 (iv).
\end{proof}
\subsection{Proof of Theorem \ref{thm:contr_Tarun}}
\begin{proof}
Using  \eqref{eq:lap_t_den} and Lemma 1, following may be stated :
\begin{align*}
	\mathcal{L} \mathbf{1}^{k-1}_n = \mathbf{0}  & \Rightarrow \left( \mathcal{D - A} \right) \mathbf{1}_n^{k-1} = \mathbf{0} \\ 
	& = \left( \mathcal{D }  \mathbf{1}_n^{k-2} - \mathcal{A} \mathbf{1}_n^{k-2} \right) \mathbf{1}_n = \mathbf{0}
\end{align*}
As $\mathcal{D}$ is a diagonal tensor of order $k$, the first term $\mathcal{D} \mathbf{1}_n^{k-2} $ is a contracted diagonal tensor of order $= k - (k-2) = 2$. The diagonal elements in the resulting $2^{nd}$ order tensor (a matrix) will be same as the diagonal of original $k$ mode tensor. 

As $\mathcal{A} $ is a \emph{super symmetrical } tensor of order $k$, the second term $\mathcal{A} \mathbf{1}_n^{k-2}$ will be a $2-$ dimensional \emph{symmetric} tensor. This is due to the fact that all $(k-2)$ modes of tensor are scaled and contracted with same vector of ones.  As all the diagonal terms for $\mathcal{A} \mathbf{1}_n^{k-2}$ are zeros,  the vertex degree can be seen to be preserved from $\mathbf{L}_T = \mathcal{D }  \mathbf{1}_n^{k-2} - \mathcal{A} \mathbf{1}_n^{k-2}$. The Laplacian of the corresponding reduced graph is described by 
\begin{align*}
	\mathbf{L}_T = \mathcal{D }  \mathbf{1}_n^{k-2} - \mathcal{A} \mathbf{1}_n^{k-2} = \mathcal{L} \mathbf{1}_n^{k-2} 
\end{align*}
\end{proof}

\subsection{Proof of Corollary \ref{corr:contr_cliq}}
The Laplacian corresponding to clique expansion \citep{paper:agarwal_2006} of a k-uniform hypergraph, denoted by $\mathbf{L_c}$ can be derived from the hypergraph tensor Laplacian as follows:
\begin{align}
    \mathbf{L_c} = (k-1) \mathcal{L} \mathbf{1}_n^{k-2} = (k-1) \mathbf{L}_T 
	\label{eq:cliq_ten_g}
\end{align}
\begin{proof}
This can be seen as a direct consequence of Lemma 1 which proposes an approach to preserve node degree. In basic clique expansion algorithm, the vertex degree is not preserved. Each node in a hyperedge of $k$ nodes will be paired up with remaining $(k-1)$ nodes. So, the same scaling factor appears in Eq. \eqref{eq:cliq_ten_g}. The proof is trivial because the input is a $k$-\emph{uniform} hypergraph.
\end{proof}
\subsection{Proof of Theorem \ref{thm:Rcut_teig}}
\begin{proof}
Given a partition of $p$ disjoint sets $\{C_1, C_2,\ldots,C_p \}$, define the $p$ indicator variables $f_j = (f_{1,j},f_{2,j}, \ldots, f_{n,j})^{'}$ defined as
\begin{align}
f_{i,j}= 
\begin{cases}
\frac{1}{\sqrt{|C_j|}} &\quad v_i \in C_j \\
0 & \text{otherwise}
\end{cases}
\end{align}
where $i \in [n]$ and $j \in [p]$. 

For any partition $C_i$, we compute $\mathcal{L} \mathbf{f}_i^{k}$
\begin{align}
    \mathcal{L} \mathbf{f}_i^{k} = \sum_{i_k = 1}^{n} \ldots \sum_{i_2 = 1}^{n} \sum_{i_1 = 1}^{n} l_{i_1i_2\ldots i_k} x_{i_1} x_{i_2} \ldots x_{i_k}
\end{align}
We use Theorem \ref{thm:Lxk_comp} to compute the above term 
\begin{align}
	\mathcal{L} \mathbf{f}_i^{k} &= \sum_{e_j \in E} w_{e_j} \left( \sum_{i_k \in e_j } f^k_{i_k} - k \prod_{i_k \in e_j} f_{i_k} \right)
\end{align}
There can be three case for each hyperedge.
\begin{enumerate}
    \item All the nodes in a hyperedge $e_j$ are assigned as $\frac{1}{|C_j|^{1/2}}$. Both the terms will be $k \frac{1}{|C_i|^{k/2}}$ and the overall term reduces to 0.
    \item All the nodes in hyperedge $e_j$ are assigned 0: Both the terms will be zero and overall term will be zero.
    \item Some of the nodes are assigned $\frac{1}{\sqrt{|C_j|}} $. The second term will be zero and the first term will reduce to $|C_i \cap e_j|  \frac{w_{e_j} }{|C|^{k/2}}$.
\end{enumerate}
So the overall term reduce to 
\begin{align}
	\mathcal{L} \mathbf{f}_i^{k} = |C_i \cap e_j|  \frac{w_{e_j} }{|C_i|^{k/2}}
\end{align}
Similarly adding other partitons, we arrive at 
\begin{align}
    \sum_{i=1}^{p}	\mathcal{L} \mathbf{f}_i^{k} =  \sum_{i=1}^{p} |C_i \cap e_j|  \frac{w_{e_j} }{|C_i|^{k/2}}
\end{align}
The RHS term in the above equation is same as the defined ratio cut for hypergraphs. It should be noted that $\mathbf{f_i}^T \mathbf{f_i} = 1$. As the objective function and constraint are same under relaxation, both the problems are equivalent and the solution can be derived from tensor Eigenvalue decompostion.
\end{proof}
\subsection{Proof of Theorem \ref{thm:Lxk_comp}}
\begin{proof}
\begin{align}
	\mathcal{L} \mathbf{x}^{k} &= \sum_{i_k = 1}^{n} \ldots \sum_{i_2 = 1}^{n} \sum_{i_1 = 1}^{n} (d_{i_1i_2\ldots i_k} - a_{i_1i_2\ldots i_k}) x_{i_1} x_{i_2} \ldots x_{i_k}  	\nonumber \\
	& = \sum_{i = 1}^{n}  d(v_i)  x_i^k - \sum_{i_k = 1}^{n} \ldots \sum_{i_2 = 1}^{n} \sum_{i_1 = 1}^{n} a_{i_1i_2\ldots i_k} x_{i_1} x_{i_2} \ldots x_{i_k} \nonumber \\
	& = \sum_{i = 1}^{n} \sum_{i_k = 1}^{n} \ldots \sum_{i_3 = 1}^{n} \sum_{i_2 = 1}^{n} a_{ii_2i_3\ldots i_k}  x_i^k - \sum_{i_k = 1}^{n} \ldots \sum_{i_2 = 1}^{n} \sum_{i_1 = 1}^{n} a_{i_1i_2\ldots i_k} x_{i_1} x_{i_2} \ldots x_{i_k} \nonumber \\
	&=  \sum_{i = 1}^{n} \left( \sum_{ (i_2, i_3, \ldots, i_k) \in e_j} \frac{w_{e_j} }{(k-1)!}  x_i^k  \right) - \left( \sum_{ (i_1, i_2, \ldots, i_k) \in e_j} \frac{w_{e_j} }{(k-1)!}  x_{i_1} x_{i_2} \ldots x_{i_k} \right) 
\end{align}
As there are $(k-1)!$ and $k!$ permutations of the first and second term respectively :
\begin{align}
	\mathcal{L} \mathbf{x}^{k} &= \sum_{e_j \in E} w_{e_j} \left(\sum_{i_t \in e_j } \frac{(k-1)!}{(k-1)!}x^k_{i_t} -  \frac{ k! }{(k-1)!}  x_{i_1} x_{i_2} \ldots x_{i_k}  \right) \nonumber \\
	&= \sum_{e_j \in E} w_{e_j} \left( \sum_{i_t \in e_j } x^k_{i_t}  - k \prod_{i_t \in e_j} x_{i_t} \right) \nonumber \\
	&= \sum_{e_j \in E} w_{e_j} \left(k \frac{\sum_{i_t \in e_j } x^k_{i_t} }{k}  - k \left( \prod_{i_t \in e_j} |x_{i_t}|^k \right)^\frac{1}{k}  (-1)^{n_s} \right) \nonumber \\ 
&=	\sum_{e_j \in E} w_{e_j} k \left( \underset{ i_t \in e_j }{AM\left(x^k_{i_t} \right)} -  \underset{ i_t \in e_j }{GM\left(|x_{i_t}|^k \right)} (-1)^{n_s} \right) \label{eq:am_gm_proof}
\end{align}
\end{proof}
\subsection{Proof of Corollary \ref{cor:normcut_teneig}}
The proof of this corollary is very similar  to the proof of Theorem \ref{thm:Rcut_teig}. In this case, we choose the indicator variable as 
\begin{align}
f_{i,j}= 
\begin{cases}
\frac{1}{\sqrt{\text{vol}(C_j)}} &\quad v_i \in C_j \\
0 & \text{otherwise}
\end{cases}
\end{align}
The next step is to compute the $\mathscr{L}\mathbf{x}^{k}$, where the normalized Laplacian tensor $\mathscr{L}$ is defined in Eq. \ref{eq:Lt_ent_norm}. Rest of the proof is very similar to the proof of Theorem \ref{thm:Rcut_teig}.
\subsection{Proof of Theorem \ref{thm:lambda_bound}}
\begin{proof}
Let $\mathbf{x}$ be $n \times 1$ vector with $x_{i_j} \in \left\{ \frac{\sqrt[k]{d_{i_j}}}{\omega}, \frac{-\sqrt[k]{d_{i_j}}}{\omega} \right\} $, where $\omega $ is defined as 
\begin{align}
\omega =\left( \sum_{i_j=1}^{n} d_{i_j}^{\frac{2}{k}} \right)^\frac{1}{2} \nonumber
\end{align}
It can be easily verified that $\mathbf{x}^T \mathbf{x} = 1$. Substituting $\mathbf{x}$ in the expression for normalized hypergraph Laplacian defined in Eq. \eqref{eq:Lt_ent_norm}. Please note that we do not use the signs of elements in the Fiedler vector to compute partitions, as discussed in the main manuscript.
\begin{align}
	\lambda_1  \leq  \mathscr{L} \mathbf{x}^m & = \sum_{e_j \in E} w_{e_j} \left( \sum_{i_j \in e_j } \frac{x^k_{i_j}}{d_{i_j}} - k \prod_{i_j \in e_j} \frac{x_{i_j}}{\sqrt[k]{d_{i_j}}} \right) \nonumber \\
					&= \sum_{e_j \in E} w_{e_j} \left(\frac{k - n_s}{\omega^k} + \frac{n_s (-1)^k}{\omega^{k}} - k (-1)^{n_s} \frac{1}{\omega^k}\right)
\end{align}
where $n_s = |\{ i_k : x_{i_k} < 0 \}|$. For even order hypergraphs, the above can be reduced to 
\begin{align}
	\lambda_1  & \leq \sum_{e_j \in E} w_{e_j} \left( \frac{k}{\omega^k} - k (-1)^{n_s} \frac{1}{\omega^k} \right) \\
 & \leq \sum_{e_j \in {\partial S}} w_{e_j} \left( \frac{2k}{\omega^k}  \right) = \sum_{e_j \in {\partial S}} w_{e_j}\left( \frac{2k}{\left( \sum_{i=1}^{n} d_i^{\frac{2}{k}} \right)^\frac{k}{2}}  \right)  \nonumber \\
&\leq \sum_{e_j \in {\partial S}} w_{e_j} \left( \frac{2k}{\left( \sum_{i=1}^{n} d_i \right)^{\frac{2}{k} \times \frac{k}{2}}} \right)  \nonumber \\
& \leq  \sum_{e_j \in {\partial S}} w_{e_j} \frac{2k}{2 \min(\text{vol}(C), \text{vol}(\bar{C}))} = k \phi(G) 
\end{align}
\end{proof}

%% file: sections/appen_exmpl.tex
\section{Examples \& Numerical Details of Experiments}
\label{sec:app_num}
\subsection{Hypergraph reduction to same graph}
\label{sec:appn_redsameGr}
Various hypergraph reduction methods have been summarized in \citep{paper:agarwal_2006}. One of the prevalent approaches is 
\begin{align}
    \mathbf{A}_r = \mathbf{H W} \mathbf{H}^T - \mathbf{D} \label{eq:inci_a}
\end{align}
It should be noticed that these reduction methods are a non-unique mapping from hypergraph to adjacency matrix. This implies that there could be multiple different hypergraphs which reduce to same graph. For example, the clique reduction approach reduces the four-uniform hypergraph  and the three-uniform hypergraph hypergraph shown in following figure to the same graph. 
\begin{figure}[H]
    \centering
    \input{figs/hyred_sameGr.tex}
    \captionsetup{justification=centering}
    \caption{Two Hypergraphs reducing to same Graph}
    \label{fig:2hy_sameGr}
\end{figure}
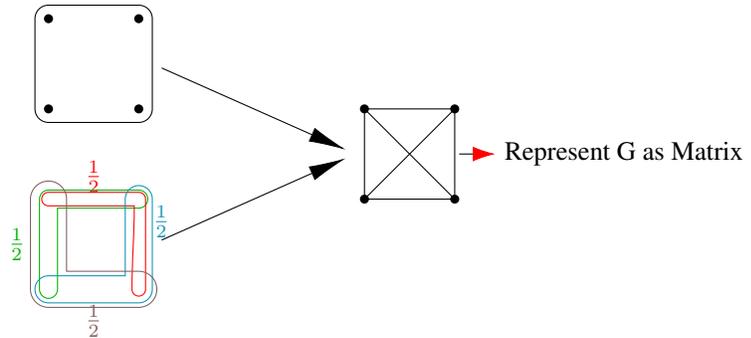
This non-uniqueness property of hypergraph reduction method plays a very crucial role in the task of hypergraph partitioning. The reduced hypergraph has lost the information about the original hypergraph structure. So there is no assurance of any analysis on reduced hypergraph to deliver correct results for the original hypergraph. To avoid the loss of information in the reduction step, we utilize the tensor-based representation of hypergraphs as shown in the next example. 

\subsection{Representation of Hypergraphs}
\label{app:def_hyp_ex}
Consider a $4-$uniform hypergraph $G(V,E)$ as shown below, 
where the set of vertices and hyperedges are defined by 
 \begin{align*}
 	V &= \{ 1, 2 ,3, 4, 5 \} \\
 	E &= \{ \{1,2,3,4\},  \{2,3,4, 5\} , \{1,2,3, 5\} \}
 \end{align*} 
 The adjacency tensor for the above hypergraph is denoted by $\mathcal{A}$ and has dimensions $5 \times 5 \times 5 \times 5$. It should be noted that the cardinality ($k$) of all $3$ hyperedges is 4 . The elements of $\mathcal{A}$ are denoted by $a_{i_1, i_2, i_3, i_4}$, where $1 \leq i_k \leq 5$. It should be noted that $\mathcal{A}$ contains $n^k = 5^4$ elements but only $m \times k! = 3 \times 4! = 72$ elements have non-zero entries. The elements corresponding to first hyperedge are described by 
 \begin{align*}
 	& a_{1234} = a_{1243} = a_{1324} = a_{1342} = a_{1423} = a_{1432} \\
 	= & a_{2134} = a_{2143} = a_{2314} = a_{2341} = a_{2413} = a_{2431} \\
 	= & a_{3214} = a_{3241} = a_{3124} = a_{3142} = a_{3421} = a_{3412} \\
 	= & a_{4231} = a_{4213} = a_{4321} = a_{4312} = a_{4123} = a_{4132} = c\\
 \end{align*}
 where $c = \frac{1}{(k-1)!} = \frac{1}{6}$. The vertex degrees can be stored in a tensor of dimension $5 \times 5 \times 5 \times 5$ with its diagonal elements being $d(v) = \begin{bmatrix}
 2 & 3 & 3 & 2 & 2
 \end{bmatrix}$.  The tensor Laplacian has dimension of $5 \times 5 \times 5 \times 5$ and its entries are given by: 
\begin{align}
l_{i_1 i_2i_3 i_4} = 
\begin{cases}
-\frac{1}{6} & \text{if $(i_1, i_2, i_3, i_4) = \{e_j\}, \quad j= \{1,2,3\}$} \\
d(v_i) & \text{if $i_1 = i_2 = i_3 = i_4 = i $} \\ 
0 & \text{otherwise}
\end{cases}
\label{eq:L_ex1}
\end{align} 
This example shows the procedure to construct the adjacency and Laplacian tensor for any $k$-uniform hypergraph. 

\subsection{Partition Cost}
\label{app:partcost}
First we show the cut cost derived from graphical. The hypergraph is shown in Figure \ref{fig:3uni_r1}. The partitions are given to be $C_1 = \{1,2,3\}$, $C_2 = \{4\}$ and $C_3 = \{5\}$. The hyperedges denoted by $e_2$ and $e_3$ have to be removed for such partition, so $\partial E = \{e_2, e_3\}$. The cost for partitions can be derived as shown below :
\begin{align}
	w_h(C_1) & =  \sum_{ e_j \in \partial E} |C_1 \cap e_j| w_{e_j} \nonumber \\
	& = |C_1 \cap e_2| w_{e_2}	+  |C_1 \cap e_3| w_{e_3} \nonumber \\
	& = 2 w_{e_2} + w_{e_3}
\end{align}
The cost of other partitions can be derived in similar fashion and are observed to be $w_h(C_2) = w_2 +  w_3$ and $w_h(C_3) = w_3$. 

Further, we compute the cut cost for reduced hypergraph. The edges between node $i$ and $j$ is named as $e_{ij}$ and the corresponding weight is denoted by $w_{e_{ij}}$.
The edges denoted by $e_{24}, e_{34}, e_{35}, e_{45}$ have to be removed for arriving at the desired partition, So $\partial E_g = \{e_{24}, e_{34}, e_{35}, e_{45}\}$ . The cut cost for such partitioning is given by:
\begin{align}
	w_g(C_1) & =  \sum_{ e_{ij} \in \partial E_g} |C_1 \cap e_{ij}| w_{e_{ij}} \nonumber \\
	&= |C_1 \cap e_{24}| w_{e_{24}} + |C_1 \cap e_{34}| w_{e_{34}} + |C_1 \cap e_{35}| w_{e_{35}} + |C_1 \cap e_{45}| w_{e_{45}} \nonumber \\
	& = w_{e_{24}} + w_{e_{34}} + w_{e_{35}} + 0 \nonumber \\
	&= w_{e_2} + (w_{e_2}+ w_{e_3}) + w_{e_3} = 2(w_{e_2} + w_{e_3})
\end{align}
The cut cost for other partitions can be calculated as $w_g(C_2) =  2(w_{e_2} + w_{e_3})$ and $w_g(C_3) =  2w_{e_3}$. 

It can be easily noticed that the cut cost for both the cases are not equal. 
On further inspection, we infer $w_g(C_i) = 2w_h(C_i)$ for $i=\{2,3\}$, which means the cut cost for partitions $C_2$ and $C_3$ in reduced hypergraph are just a scaled version of costs involved in original hypergraph. The same relation doesn't hold for partition $C_1$ due to the presence of the term $|C_i \cap e_j|$.

\subsection{Numerical details of Example \ref{ex:3uni_2part}}
The Fiedler eigenvalue for this hypergraph is reported to be $0.0372$ and the corresponding eigenvector is given by:
\begin{align}
    \mathbf{f} = \begin{bmatrix}
     0.34 &
    0.21 &
    0.18 &
    0.11 &
   -0.07 &
    0.05 &
    0.12 &
    0.38 &
    0.39 &
    0.39 &
    0.38 &
    0.39
    \end{bmatrix}^T
\end{align}
It is clear that the classical spectral approach does not produce the optimal partitions and hence we compute the hyperedge scores as prescribed by the proposed algorithm. 
\begin{table}[H]
\centering
\begin{tabular}{c|c}
\toprule
Hyperedge & Score \\
\midrule 
   $\{ 1 ,     2    ,  3 \}$ &     0.0170 \\
   $\{ 4  ,    5    ,  6  \}$ &    0.0060 \\
   $\{ 2  ,    3,      4  \}$ &    0.0041 \\ 
   $\{ 3  ,    4 ,     7  \}$ &    0.0036 \\
   $\{ 1  ,    8 ,    11  \}$ &    0.0033 \\
   $\{ 4  ,    6 ,     7  \}$ &    0.0024 \\
   $\{ 8  ,    9 ,    10 \}$ &     0.0005 \\
   $\{10  ,   11 ,    12  \}$ &    0.0005 \\
   $\{ 9  ,   10 ,    12 \}$ &     0  \\
\bottomrule
\end{tabular}
\captionsetup{justification=centering}
\caption{Hyperedge-Score for Example \ref{ex:3uni_2part}}
\label{tab:hysc_3uni_2part}
\end{table}
We observe the maximum score of $0.017$ for the hyperedge $\{1,2,3\}$ and hence cut it to obtain the optimal partitions $A_1 = \{2, 3, 4, 5, 6, 7\}$ and $\bar{A}_1$.
\subsection{ Numerical details of Example \ref{ex:roach}}
\label{sec:app_roach}
Consider the cockroach graph shown in Figure \ref{fig:roach} for $t=3$ and hence 12 nodes. The Fiedler vector for this graph is given by:
{\small
\begin{align}
    \mathbf{f} = \begin{bmatrix}
    -0.49 & -0.41 &  -0.26 &   -0.07 &   -0.02 &    -0.01 &   0.49 & 0.41 &  0.26 &   0.07 &   0.02 &    0.01 
    \end{bmatrix}^T
\end{align}}
So the partitions defined based on the sign of elements in $\mathbf{v}$ are $A_1 = \{v_1,v_2,v_3,v_4,v_5,v_6\}$ and $A_2 = \bar{A_1}$. So, RatioCut$(A_1, \bar{A_1}) = \frac{3}{6} + \frac{3}{6} = 1$. 

The next step is to apply the proposed algorithm. The edge score computed from the proposed algorithm are:
\begin{table}[H]
\centering
\begin{tabular}{c|c}
\toprule
Edge & Score \\
\midrule 
    $\{v_3,v_4\}$ &     0.0371\\
    $\{v_9,v_{10}\}$ &   0.0371\\
    $\{v_4,v_{10}\}$  &    0.0228\\
    $\{v_2,v_3\}$  &    0.0228\\
    $\{v_8,v_9\}$  &   0.0222\\
    $\{v_1, v_2\}$  &   0.0222\\
    $\{v_7, v_8\}$  &   0.0066\\
    $\{v_4,v_5\}$  &   0.0066\\
    $\{v_{10},v_{11}\}$  &   0.0029\\
    $\{v_{5}, v_{11}\}$  &   0.0029\\
    $\{v_{6},  v_{12}\}$  &   0.0002\\
     $\{v_{11},  v_{12}\}$  &   0.0002\\
     $\{v_{5},  v_{6}\}$  &   0.0002\\
\bottomrule
\end{tabular}
\captionsetup{justification=centering}
\caption{Edge-Score for  \ref{ex:3uni_2part}}
\label{tab-end}
\end{table}
The edges $\{v_3,v_4\}$ and $\{v_9, v_{10}\}$ are removed as they are of  maximum edge score of $0.0371$. So the partitions are $B_1 = \{v_1,v_2,v_3,v_7,v_8, v_9\}, B_2 = \bar{B_1}$. Therefore, RatioCut$(B_1, \bar{B_1}) = \frac{2}{6} + \frac{2}{6} = 0.66$. It can be clearly observed that the partitions obtained from the proposed algorithm have a lower RatioCut value compared to the existing method. 

We observe the superior performance of the proposed algorithm for $t = \{3,4,\ldots,20\}$.

%% file: figs/hyred_sameGr.tex
\begin{tikzpicture}[scale = 0.6]
    \node (v1) at (0,2) {};
    \node (v2) at (2,2) {};
    \node (v3) at (0,0) {};
    \node (v4) at (2,0) {};

    \draw[] ($(v2) + (0.3,0)$) 
        to[out=270,in=90] ($(v4) + (0.3,0)$)
        to[out=270,in=0] ($(v4) + (0,-0.3)$)
        to[out=180,in=0] ($(v3) + (0,-0.3)$)
        to[out=180,in=270] ($(v3) + (-0.3,0)$)
        to[out=90,in=270] ($(v1) + (-0.3,0)$)
        to[out=90,in=180] ($(v1) + (0,0.3)$)
        to[out=0,in=180] ($(v2) + (0,0.3)$)
        to[out=0,in=90] ($(v2) + (0.3,0)$);
  
  \foreach \v in {1,2,...,4} {
        \fill (v\v) circle (0.1);
    }
    
  
    \node (r1) at (7,0) {};
    \node (r2) at (9,0) {};
    \node (r3) at (7,-2) {};
    \node (r4) at (9,-2) {};
    
    \foreach \v in {1,2,...,4} {
        \fill (r\v) circle (0.1);
    }
    

   \draw[] ($(r1)$) to ($(r2)$);
   \draw[] ($(r1)$) to ($(r3)$);
   \draw[] ($(r1)$) to ($(r4)$);
   \draw[] ($(r2)$) to ($(r3)$);
   \draw[] ($(r2)$) to ($(r4)$);
   \draw[] ($(r3)$) to ($(r4)$);
   
   \node (v5) at (2.3,1) {};
   \node (r5) at (6.8,-1) {};
   \draw[-{Latex[length=5mm, width=2mm]}] (v5)--(r5);
   \draw[-{Latex[red,length=3mm, width=2mm]}] (9.1,-1)--(9.9,-1) node[right]{Represent G as Matrix};

    \node (g1) at (0,-2) {};
    \node (g2) at (2,-2) {};
    \node (g3) at (0,-4) {};
    \node (g4) at (2,-4) {};

    \node (e1) at (1,-1.5) {\textcolor{red}{$\frac{1}{2}$}};
    \node (e1) at (2.5,-2.5) {\textcolor{black!30!cyan}{$\frac{1}{2}$}};
    \node (e1) at (1,-4.7) {\textcolor{black!50!pink}{$\frac{1}{2}$}};
    \node (e1) at (-0.7,-3) {\textcolor{black!30!green}{$\frac{1}{2}$}};
    
     \draw[black!30!green] ($(g1)+(-0.2,0)$) 
        to[out=270,in=90] ($(g3) + (-0.2,0)$) 
        to[out=270,in=180] ($(g3) + (0,-0.2)$)
        to[out=0,in=270] ($(g3) + (0.2,0)$)
        to[out=90,in=270] ($(g1) + (0.2,-0.2)$)
        to[out=0,in=180] ($(g2) + (0,-0.2)$)
        to[out=0,in=270] ($(g2) + (0.2,0)$)
        to[out=90,in=0] ($(g2) + (0,0.2)$) 
        to[out=180,in=0] ($(g1) + (0,0.2)$)
        to[out=180,in=90] ($(g1) + (-0.2,0)$) ;

    \draw[red] ($(g1)+(0,-0.15)$) 
        to[out=0,in=180] ($(g2) + (-0.1,-0.15)$) 
        to[out=270,in=90] ($(g4) + (-0.15,0)$)
        to[out=270,in=180] ($(g4) + (0,-0.15)$)
        to[out=0,in=270] ($(g4) + (0.15,0)$)
        to[out=90,in=270] ($(g2) + (0.15,0)$)
        to[out=90,in=0] ($(g2) + (0,0.15)$)
        to[out=180,in=0] ($(g1) + (0,0.15)$)
        to[out=180,in=90] ($(g1) + (-0.15,0)$)
        to[out=270,in=180] ($(g1) + (0,-0.15)$);
    
    \draw[black!30!cyan] ($(g2) + (0.3,0)$) 
        to[out=270,in=90] ($(g4) + (0.3,0)$)
        to[out=270,in=0] ($(g4) + (0,-0.3)$)
        to[out=180,in=0] ($(g3) + (0,-0.3)$)
        to[out=180,in=270] ($(g3) + (-0.3,0)$)
        to[out=90,in=180] ($(g3) + (0,0.3)$)
        to[out=0,in=180] ($(g4) + (-0.3,0.3)$)
        to[out=90,in=270] ($(g2) + (-0.3,0)$)
        to[out=90,in=180] ($(g2) + (0,0.3)$)
        to[out=0,in=90] ($(g2) + (0.3,0)$);
    
    \draw[black!50!pink] ($(g4) + (0.4,0)$)
        to[out=270,in=0] ($(g4) + (0,-0.4)$)
        to[out=180,in=0] ($(g3) + (0,-0.4)$)
        to[out=180,in=270] ($(g3) + (-0.4,0)$)
        to[out=90,in=270] ($(g1) + (-0.4,0)$)
        to[out=90,in=180] ($(g1) + (0,0.4)$)
        to[out=0,in=90] ($(g1) + (0.4,0)$)
        to[out=270,in=90] ($(g3) + (0.4,0.4)$)
        to[out=0,in=180] ($(g4) + (0,0.4)$)
        to[out=0,in=90] ($(g4) + (0.4,0)$);


    \node (g5) at (2.3,-3) {};
   \draw[-{Latex[length=5mm, width=2mm]}] (g5)--(r5);
    
\end{tikzpicture} 